\documentclass[letterpaper,twocolumn,fleqn]{article} 

\usepackage{ist}
\usepackage{times}
\usepackage{epsfig}
\usepackage{graphicx}
\usepackage{subcaption}
\usepackage{amsmath}
\usepackage{amssymb}
\usepackage{bbold}
\usepackage{soul}
\usepackage{float}
\usepackage[pagebackref=true,breaklinks=true,letterpaper=true,colorlinks,bookmarks=false]{hyperref}

\title{Holistic Image Manipulation Detection using Pixel Co-occurrence Matrices}

\author{Lakshmanan Nataraj\textsuperscript{1},
Michael Goebel\textsuperscript{2}, 
Tajuddin Manhar Mohammed\textsuperscript{1}, 
Shivkumar Chandrasekaran\textsuperscript{1,2}, 
B. S. Manjunath\textsuperscript{1,2}
\\
\textsuperscript{1}Mayachitra Inc; Santa Barbara, CA \\
\textsuperscript{2}University of California, Santa Barbara}

\date{} 
\begin{document} 

\maketitle 

\thispagestyle{empty} 


\begin{abstract}
Digital image forensics aims to detect images that have been digitally manipulated.
Realistic image forgeries involve a combination of splicing, resampling, region removal, smoothing and other manipulation methods. 
While most detection methods in literature focus on detecting a particular type of manipulation, it is challenging to identify doctored images that involve a host of manipulations.
In this paper, we propose a novel approach to holistically detect tampered images using a combination of pixel co-occurrence matrices and deep learning.
We extract horizontal and vertical co-occurrence matrices on three color channels in the pixel domain and train a model using a deep convolutional neural network (CNN) framework.
Our method is agnostic to the type of manipulation and classifies an image as tampered or untampered.
We train and validate our model on a dataset of more than 86,000 images. Experimental results show that our approach is promising and achieves more than 0.99 area under the curve (AUC) evaluation metric on the training and validation subsets. 
Further, our approach also generalizes well and achieves around 0.81 AUC on an unseen test dataset comprising more than 19,740 images released as part of the Media Forensics Challenge (MFC) 2020. 
Our score was highest among all other teams that participated in the challenge, at the time of announcement of the challenge results. 
\end{abstract}


\section{Introduction}
\label{sec:intro}

Fake images and videos are ubiquitous in today's world.
These fake media are becoming a growing threat to information reliability. 
With the availability of powerful apps and tools, as well as recent Artificial Intelligence (AI) based automated technologies such as Deepfakes, it has become very trivial to generate fake manipulated images.
The field of Digital image forensics aims to develop methods to automatically detect whether digital images have been tampered with or not. 

\begin{figure}[t]
\centering
\includegraphics[scale=0.3]{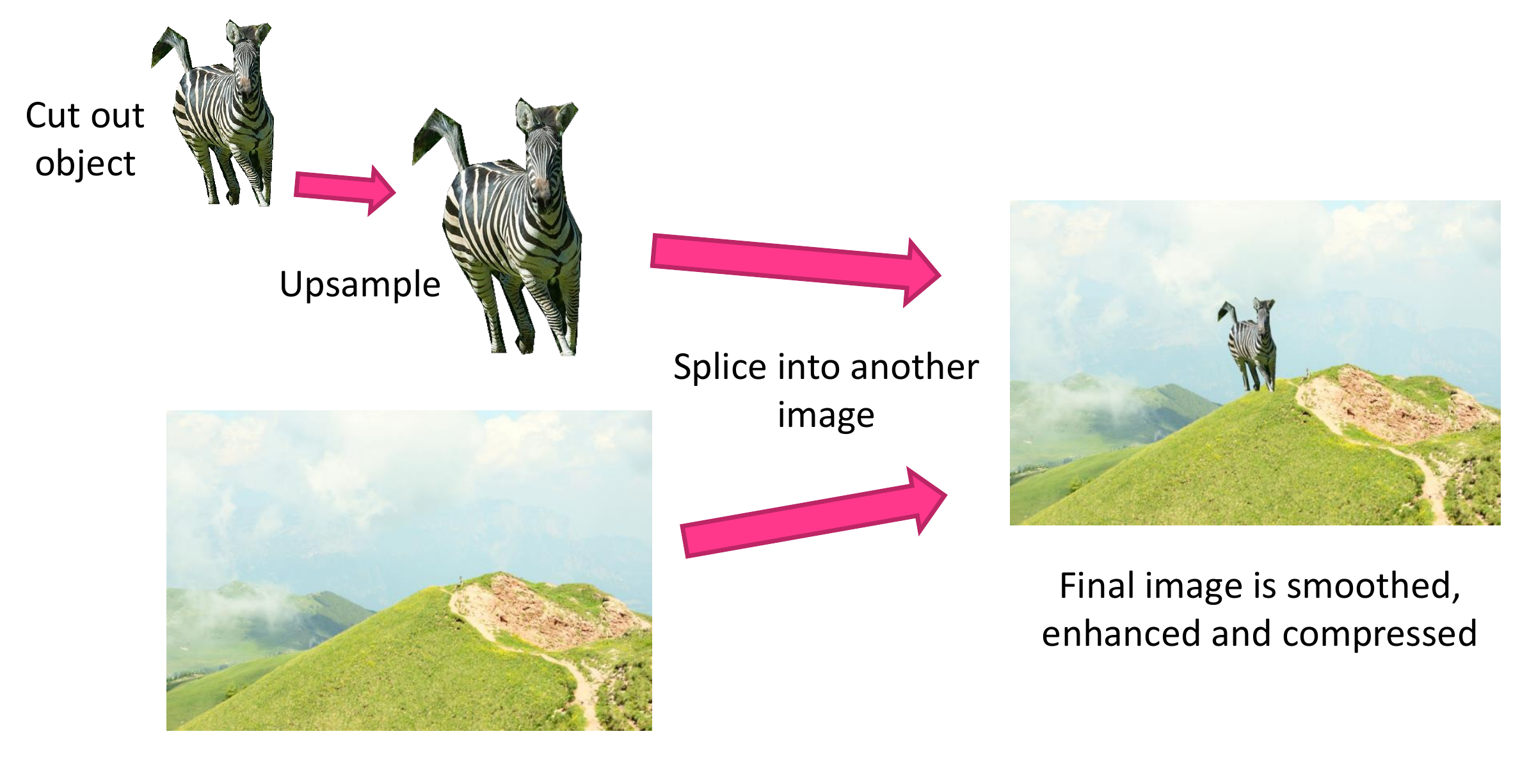}
\vspace{-5pt}
\caption{An example of realistic image forgery involving a host of manipulations.} 
\label{fig:illus}
\vspace{-10pt} 
\end{figure}

There are many types of image forgeries such as splicing objects from one image to another, removing objects or regions from images, creating copies of objects in the same image, and more. 
To detect these manipulations, researchers have proposed several methods based on a variety of techniques such as JPEG compression artifacts, resampling detection, lighting artifacts, noise inconsistencies, camera sensor noise, and many more. 
However, most of these techniques focus on a specific type of manipulation or groups of similar manipulation operations while in a realistic scenario, several manipulation operations are applied when creating doctored images.
For example, when an object is spliced onto an image, it is often accompanied by other operations such as scaling, rotation, smoothing, contrast enhancement, to name a few (Fig.~\ref{fig:illus}).
Very few studies address these challenging scenarios with the aid of Image Forensics challenges and competitions such as IEEE Image Forensics challenge~\cite{2013-ifc-challenge} and the recent Media Forensics challenges~\cite{nist2017,mfc2019,mfc2020-res}.
These competitions try to mimic a realistic scenario and contain a large number of doctored images which involve several types of image manipulations.
In order to detect these types of realistic tampered images, it is important to look into holistic image manipulation detection techniques which are not focused on detecting a single type of manipulation but can detect a host of different manipulations.
In this paper, we focus on a holistic method to determine if an image has been tampered or not.

When an image is doctored, the statistics of the pixels in the doctored image are usually different than the untampered image. 
Hence, methods that look for deviations from natural image statistics could be effective in holistic detection of tampered images.
These methods have been well studied in the field of steganalysis which aims to detect the presence of hidden data in digital images.
One such method is based on analyzing co-occurrences of pixels by computing a co-occurrence matrix.
Traditionally, this method uses hand crafted features computed on the co-occurrence matrix and a machine learning classifier such as support vector machines determines if a message is hidden in the image~\cite{sullivan2005steganalysis,sullivan2006steganalysis}.
Other techniques involve calculating image residuals or passing the image through different filters before computing the co-occurrence matrix~\cite{pevny2010steganalysis,fridrich2012rich, cozzolino2017recasting}.

Inspired by steganalysis and natural image statistics as well as recent advances in deep learning, we propose a novel method to holistically detect if images are manipulated, using a combination of pixel co-occurrence matrices and deep learning.
First, we compute horizontal and vertical co-occurrence matrices on the different color channels of an image. 
These capture the distribution of neighboring pixels horizontally and vertically.
Then we stack these matrices and pass them directly through a deep learning framework and allow the network to learn important features that can separate tampered images from untampered images.  
We also avoid computation of residuals or passing an image through various filters, but rather compute the co-occurrence matrices on the image pixels itself.
Recently, we showed that this approach is effective in detection of Generative Adversarial Network (GAN) generated images~\cite{nataraj2019detecting}.
In this paper, we extend this approach to the more general case of detection of manipulated images.
Experimental results on the evaluation dataset of the recently concluded Media Forensics Challenge (MFC) 2020 showed that our approach is promising and a viable method for holistic image manipulation detection. 
Our approach was the top performing approach in the challenge~\cite{mfc2020-res} (as shown in slide 47).

\section{Related Work}
\label{sec:relwork}

Several methods have been proposed to detect digital image manipulations (see~\cite{forgery_1,forgery_2,forgery_3,forgery_4,forgery_5} for an overview).
These included detection of splicing, resampling, copy-move, object removal, JPEG compression artifacts, machine learning and deep learning techniques, to name a few. 
Many techniques have been proposed to detect resampling~\cite{popescu-farid-resampling, babak-radon, kirchner-local, feng2012normalized, ryu2014estimation},  copy-move~\cite{li2015segmentation, cozzolino2015efficient}, splicing~\cite{guillemot2014image,muhammad2014image}, seam carving~\cite{sarkar2009detection,liu2015improved} and inpainting based object removal~\cite{wu2008detection,liang2015efficient} or artifacts arising from artificial intelligence (AI) generated images~\cite{marra2018detection,nataraj2019detecting,zhang2019detecting,goebel2020detection,barni2020cnn}.
Several approaches exploit JPEG blocking artifacts to detect tampered regions~\cite{farid2009exposing,luo2010jpeg,bianchi2011improved}. 
Recent efforts in detecting manipulations exploit deep learning based approaches~\cite{bayar2016deep, bayar2017design,rao2016deep,chen2018focus,zhou2018learning,bunk2017detection,bappy2019hybrid}.
These include detection of generic manipulations~\cite{bayar2016deep, bayar2017design}, resampling~\cite{bayar2017resampling}, splicing~\cite{rao2016deep}, reflection~\cite{wengrowski2017reflection,sun2017object} and bootleg~\cite{buccoli2014unsupervised}.
Unlike most of these works, our approach combines classical statistical approaches with modern deep learning techniques by passing co-occurrence matrices through a deep learning framework, and then holistically detecting whether an image has been manipulated or not.
In prior work~\cite{nataraj2019detecting}, we had shown that this approach was effective in detecting images which were generated using a GAN.
In this paper, we extend this approach to the more general case of holistic detection of manipulated images.
Further, we compute both horizontal and vertical co-occurrence matrices and pass a 6-D tensor through a deep learning framework, whereas only horizontal co-occurrence matrices are used in~\cite{nataraj2019detecting}.

\section{Methodology}

\subsubsection{Co-Occurrence Matrix Computation}
The first step is computing pixel co-occurrence matrices on different image channels and then passing these stacked matrices through a deep learning network.
The co-occurrence matrices represent a two-dimensional histogram of pixel pair values in a region of interest.
The vertical axis of the histogram represents the first value of the pair, and the horizontal axis, the second value.
Eqn.~\ref{eq:co_mtx_comp} shows an example of this computation for a vertical pair.

\begin{equation}
    C_{i,j} = \sum_{m,n} \begin{cases} 1, \ \  I[m,n] = i \: \ and \: I[m+1,n] = j \\ 0, \ \ otherwise
    \end{cases}
    \label{eq:co_mtx_comp}
\end{equation}

\begin{figure}[t]
\centering
\captionsetup{justification=centering}
\includegraphics[scale=0.8]{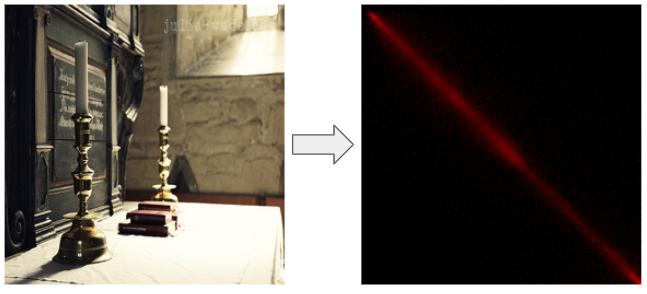}
\vspace{4pt}
\caption{Example of an image and it's pixel co-occurrence matrix computed on an image channel} 
\label{fig:coocc-eg}
\vspace{-10pt}
\end{figure}

Under the assumption of 8-bit pixel depth (grayscale images), this will always produce a co-occurrence matrix of size 256x256 for a single image channel. 
This is a key advantage of using co-occurrence matrices since it will allow for the same network to be trained and tested on a variety of images without resizing.
Fig.~\ref{fig:coocc-eg} shows an example of an image and a co-occurrence matrix computed on an image channel.  
In this paper, we compute horizontal and vertical co-occurrence matrices on the red, green and blue channels of an image.

\begin{figure*}[t]
\centering
\captionsetup{justification=centering}
\includegraphics[scale=0.5]{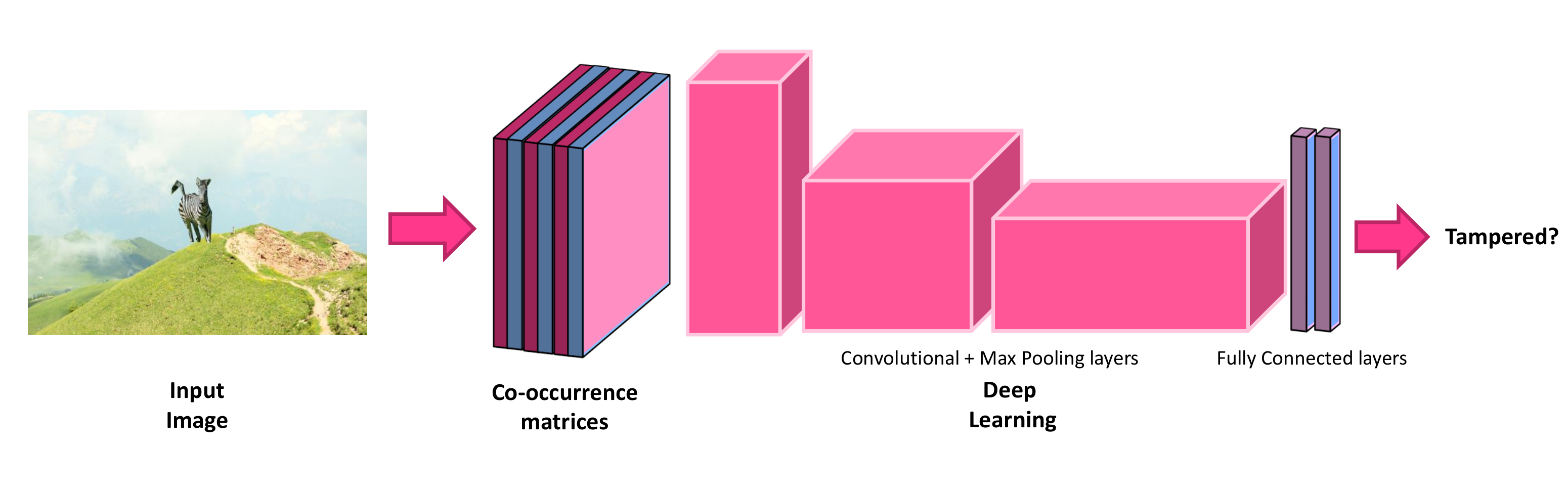}
\vspace{-4pt}
\caption{An end-to-end framework to holistically detect tampered images.} 
\label{fig:ovw}
\end{figure*}

\subsubsection{End-to-end Learning}

Now, we describe our end-to-end system to holistically detect manipulations in digital images.
Fig.~\ref{fig:ovw} shows the block schematic of our system.
First step is to compute the horizontal and vertical co-occurrence matrices on the RGB channels to obtain two 3x256x256 tensors.
These are then stacked to obtain a 6x256x256 tensor.
This tensor is then passed through a multi-layer deep convolutional neural network.
We use a ResNet50 network~\cite{he2016deep} to train our model, with over 50 epochs, a batch size of 64, Adam optimizer, and cross-entropy loss.
After each of the epochs, the model was evaluated on the validation set and the weights corresponding to the lowest validation loss were saved, and used for the remainder of the tests.


\begin{table*}
\begin{center}
\begin{tabular}{|c|c|c|c|c|}
\hline
Dataset Name & Dev/Eval & Total No. of images & No. of untampered images & No. of tampered images\\\hline
MFC18-Dev1-Ver1   & Dev & 6,708 & 997 & 5,711 \\\hline
MFC18-Dev1-Ver2   & Dev & 5,579 & 1,163 & 4,416 \\\hline
MFC18-Dev1-Ver3  & Dev & 5,557 & 1,162 & 4,395 \\\hline 
MFC18-Dev2-Ver1  & Dev & 38,339 & 1,429 & 36,910 \\\hline 
MFC19-Dev1-Ver1  & Dev & 3,592 & 3462 & 130 \\\hline 
NC17-Eval-Part1  &  Eval & 4,073  & 2,665 & 1,408 \\\hline 
MFC18-Eval-Part1  & Eval &  17,406 & 14,143 & 3,263 \\\hline 
MFC19-Eval-Part1  & Eval &  16,024 & 10,276 & 5,7408 \\\hline 
MFC20-Eval-Part1  & Eval &  19,747 & 6,124 & 13,623 \\\hline 
\end{tabular}
\end{center}
\caption{Summary of datasets provided for development (Dev) and evaluation (Eval)}
\label{tab:medifor-datasets} 
\vspace{-1mm}
\end{table*}


\section{Experiments}
\label{sec:exps}

\subsection{Datasets}

We evaluate our approach on datasets provided as part of the DARPA Media Forensics (MediFor) program~\cite{2019-darpa-medifor}, in which we were one of the participating teams. 
These datasets are further divided into Development datasets and Evaluation datasets, and broadly contain two classes of images: untampered and tampered images.
The tampered images are further broken down by sub-classes depending on the manipulation operations applied on the images. 
These operations include splicing, object removal, copy-move, contrast enhancement, blurring, cropping, antiforensics, and social media laundering, to name a few. 
Later releases of the datasets also included Generative Adversarial Networks (GAN) based manipulations.
More details on the different types of manipulations and how the datasets were curated can be found in~\cite{guan2019mfc}.

\noindent \textbf{Development Datasets:} These datasets comprise various releases of tampered and untampered images between 2017 and 2019, and are mainly provided to the performers for the purpose of development of algorithms (summarized in Tab.~\ref{tab:medifor-datasets}).
We use these datasets for training and validating our models and then test them on the evaluation datasets.

\noindent \textbf{Evaluation Datasets:} These datasets were provided to the teams as part of annual challenges held between 2017 and 2020.  
The labels of the images were not released at the time of challenge but released a few months after the challenge is completed. 
Tab.~\ref{tab:medifor-datasets} summarizes the distribution of images as part of the annual challenges (tagged by Eval).

\subsection{Results on Development datasets}
\begin{figure}[t]
  \centering
  \begin{subfigure}{.24\textwidth}
  \centering
  \includegraphics[scale=0.28]{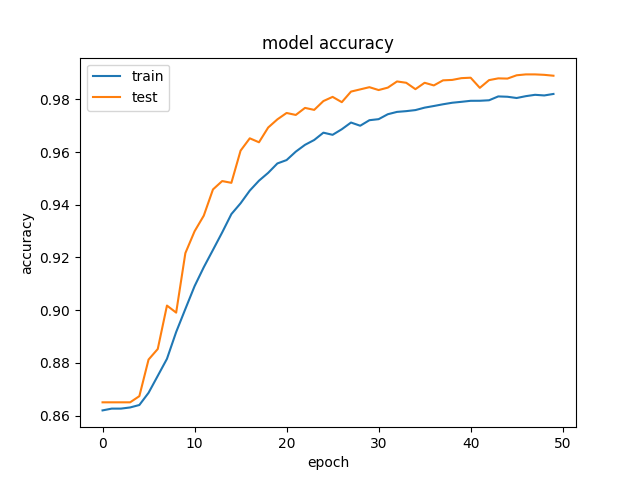}
  \caption{}
\end{subfigure}%
\begin{subfigure}{.24\textwidth}
  \centering
  \includegraphics[scale=0.28]{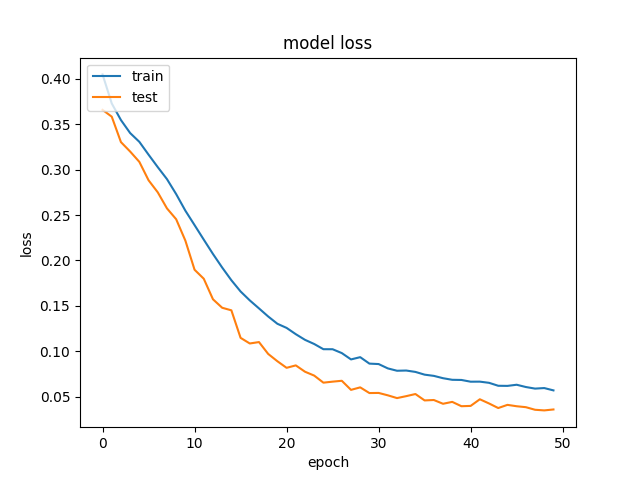}
  \caption{}
\end{subfigure}
\vspace{8pt}
\caption{Model accuracy (a) and loss (b) on Development datasets}
\label{fig:acc-loss}
\end{figure}

We combine all the datasets of the Development datasets into one dataset comprising a total of 59,774 images.
Of this, we choose 90\% for training and 10\% for validation.
Fig.~\ref{fig:acc-loss}(a,b) show the model accuracy and loss respectively for the training and validation data. 
We obtained a high training and validation accuracy, as well as low losses on both sets.

\subsection{Results on Evaluation dataset}
We evaluated our holistic detector on the MFC 2020 Evaluation dataset of 19,000+ images.
Initially, we trained a model only on the Development datasets (59,000+ images) and obtained an area under the curve (AUC) of the receiver operating characteristic (ROC) curve (metric used in the challenge) of 0.77.
Then, we trained a model that included both the Development datasets and Evaluation datasets  (37,000+ images) of previous years (whose ground truth was made available at the time of the 2020 challenge), and obtained an AUC of \textbf{0.81} as shown in Tab.~\ref{tab:eval20-res-all}.
This was the highest reported AUC at the time of announcement of the challenge results~\cite{mfc2020-res} (slide 47). 

\subsubsection{Boosting Detection Performance using Fusion}
To further boost the performance, we trained two more models, one that uses only horizontal co-occurrence matrices and the other that uses only vertical co-occurrence matrices. 
Then, we compute the average prediction values of the three models as the final prediction. 
This fusion further resulted in a slight increase in the AUC to 0.82 (Tab.~\ref{tab:eval20-res-fus}).

\begin{table}
\begin{center}
\begin{tabular}{|c|c|c|c|c|}
\hline
\textbf{Training Datasets} & \textbf{AUC-ROC} \\\hline
Dev only   & 0.77 \\\hline
Dev + Eval (2018-19)   & \textbf{0.81}  \\\hline
\end{tabular}
\end{center}
\vspace{-6pt}
\caption{Results on MFC 2020 Evaluation dataset}
\label{tab:eval20-res-all} 
\end{table}

\begin{table}
\begin{center}
\begin{tabular}{|c|c|c|c|c|}
\hline
\textbf{Co-occurrence Matrix Direction} & \textbf{AUC-ROC} \\\hline
Horizontal   & 0.80 \\\hline
Vertical   & 0.80  \\\hline
Horizontal + Vertical  & 0.81 \\\hline
Fusion  & \textbf{0.82}  \\\hline
\end{tabular}
\end{center}
\vspace{-6pt}
\caption{Improvements using Fusion}
\label{tab:eval20-res-fus} 
\end{table}

\subsubsection{Results on Selective Manipulation Types}

Next we evaluate the performance of our holistic image manipulation detector on selective manipulation types such as splicing, blurring, and others.
This analysis helps in understanding on which manipulation types our model succeeds/fails in detecting. 
Tab.~\ref{tab:eval20-res-select} summarizes the results on selective manipulations used in the MFC 2020 evaluation dataset.
We can see that our model achieves very high AUC (greater than 0.9) on manipulations such as Global Intensity Normalization and Social Media Laundering, and high AUC on other types such as Blurring, GAN and Computer Graphics (CGI) based manipulations, Resize, Antiforensics, to name a few. 
There are also manipulation types such as Splicing, Cropping for which the AUC is not high. 
This could be because the global statistics of the pixels do not change a lot in manipulations such as cropping or splicing. 
One solution to increase the performance for these manipulation types is to consider local block statistics instead of the global image statistics, which we will explore in future.

\begin{table}
\begin{center}
\begin{tabular}{|c|c|c|c|c|}
\hline
\textbf{Manipulation Type} & \textbf{AUC-ROC} \\\hline
Splice   & 0.73 \\\hline
Clone  & \textbf{0.82}  \\\hline
SpliceClone  & 0.73  \\\hline
Crop   & 0.78 \\\hline
Resize  & \textbf{0.82} \\\hline
Global Intensity Normalization   & \textbf{0.94} \\\hline
Intensity Change   & \textbf{0.81} \\\hline
Antiforensic-PRNU  &\textbf{0.83}  \\\hline
Antiforensic-CFA  & \textbf{0.81}  \\\hline
Social Media Laundering   & \textbf{0.93} \\\hline
Global Blurring  & \textbf{0.80} \\\hline
Local Blurring   & \textbf{0.83} \\\hline
GAN  & \textbf{0.83}  \\\hline
Non-GAN CGI   & \textbf{0.85} \\\hline
Distortion  & \textbf{0.89}  \\\hline
\end{tabular}
\end{center}
\caption{Results on Selective Manipulations in MFC 2020 Evaluation dataset}
\label{tab:eval20-res-select} 
\vspace{-1mm}
\end{table}

\section{Conclusion and Future Work}
\label{sec:conc}

In this paper, we presented a method to holistically detect tampered images which are usually created using many different manipulation operations. 
We combine statistical co-occurrence matrices with deep learning by extracting horizontal and vertical co-occurrence matrices on the image channels and then passing them through a deep neural network to classify if images have been tampered or not.
Our approach is holistic since it does not depend on a particular image manipulation type.
Experiments on a recently concluded media forensics challenge showed that our approach achieved promising results.
In future, we will focus on combining our holistic image manipulation detection method with methods that focus on detecting specific manipulations. 


\section{Acknowledgements}
This research was developed with funding from the Defense Advanced Research Projects Agency (DARPA). The views, opinions and/or findings expressed are those of the author and should not be interpreted as representing the official views or policies of the Department of Defense or the U.S. Government.








{\small
\bibliographystyle{spiejour}
\bibliography{coccur-forensics}
}

\section{Acknowledgements}
This research was developed with funding from the Defense Advanced Research Projects Agency (DARPA).
The views, opinions and/or findings expressed are those of the author and should not be interpreted as representing the official views or policies of the Department of Defense or the U.S. Government. 
The paper is approved for public release, distribution unlimited.

\begin{biography}

\textbf{Lakshmanan Nataraj} received his B.E degree from Sri Venkateswara College of Engineering, Anna University in 2007, and the Ph.D. degree in the Electrical and Computer Engineering from the University of California, Santa Barbara in 2015. 
He is currently a Senior Research Staff Member at Mayachitra Inc., Santa Barbara, CA. 
His research interests include malware analysis and image forensics. 

\textbf{Michael Goebel} received his B.S. and M.S. degrees in Electrical Engineering from Binghamton University in 2016 and 2017. He is currently a PhD student in Electrical Engineering at University of California Santa Barbara.

\textbf{Tajuddin Manhar Mohammed} received his B.Tech (Hons.) degree from Indian Institute of Technology (IIT), Hyderabad, India in 2015 and his M.S. degree in Electrical and Computer Engineering from University of California Santa Barbara (UCSB), Santa Barbara, CA in 2016. After obtaining his Masters degree, he obtained a job as a Research Staff Member for Mayachitra Inc., Santa Barbara, CA. His recent research efforts include developing computer vision techniques for image forensics and cyber security.

\textbf{Shivkumar Chandrasekaran} received his Ph.D. degree in Computer Science from Yale
University, New Haven, CT, in 1994.
He is a Professor in the Electrical and Computer
Engineering Department, University of California,
Santa Barbara. His
research interests are in Computational Mathematics

\textbf{B. S. Manjunath} (F’05) received the Ph.D. degree in Electrical Engineering from the University of Southern California in 1991. He is currently a Distinguished Professor at the ECE Department at the University of California at Santa Barbara. 
He has co-authored about 300 peer-reviewed articles.
His current research interests include image processing, computer vision and biomedical image analysis.

\end{biography}

\end{document}